%% file: main.tex
\documentclass[10pt]{article}
\usepackage{spconf,amsmath,graphicx}
\usepackage{subfiles}
\usepackage{color}
\usepackage{booktabs}
\usepackage{amsfonts}
\usepackage{footnote}
\usepackage{hyperref}
\usepackage{graphbox}
\usepackage{float}
\usepackage{multirow}
\usepackage{array}
\usepackage{subcaption}
\usepackage[dvipsnames]{xcolor}
\graphicspath{{figures/}{../figures/}}

\usepackage[colorinlistoftodos]{todonotes}

\newcommand{\shruti}[1]{\textcolor{blue}{\bf\small [#1 --SP]}}

\newcommand{\E}[1]{\underset{#1}{\mathbb{E}}} 
\newcommand{\tr}{\textrm{tr }} 
\newcommand{\Rm}[2]{\mathbb{R}^{#1 \times #2}} 

\newcommand{\X}{\textrm{X}}

\newcommand{\U}{\textrm{U}}
\newcommand{\V}{\textrm{V}}
\newcommand{\G}{\textrm{G}}
\newcommand{\I}{\textrm{I}}
\newcommand{\C}{\textrm{C}}

\newfont{\testpt}{ptmr at 9.19pt}

\title{exploring multiview correlations in open-domain videos}
\title{learning from multiview correlations in open-domain videos}

\name{Nils Holzenberger$^{1,*}$, Shruti Palaskar$^{2,}$\sthanks{Equal contribution},  Pranava Madhyastha$^{3}$, Florian Metze$^{2}$, Raman Arora$^{1}$}
\address{$^{1}$ Johns Hopkins University, Baltimore, MD, USA ~\\ $^{2}$ Carnegie Mellon University, Pittsburgh, PA, USA ~\\  $^{3}$ Imperial College London, UK ~\\  nholzen1@jhu.edu, spalaska@cs.cmu.edu, pranava@imperial.ac.uk, fmetze@cs.cmu.edu, arora@cs.jhu.edu}

\begin{document}
\testpt
\maketitle
\begin{abstract}
\subfile{sections/0abstract}

\end{abstract}


\vspace*{-5pt}
\section{Introduction}
\label{sec:intro}
\subfile{sections/1introduction.tex}

\vspace*{-8pt}
\section{Related Work}
\label{sec:related_work}
\subfile{sections/2related_work.tex}

\vspace*{-5pt}
\section{Methods}
\label{sec:methods}
\subfile{sections/3methods.tex}

\vspace*{-5pt}
\section{Experiments}
\label{sec:experiments}
\subfile{sections/4experiments.tex}

\vspace*{-5pt}
\section{Results and Discussions}
\label{sec:results_discussion}
\subfile{sections/5results_discussion.tex}

\vspace*{-5pt}
\section{Conclusion}
\label{sec:conclusions}
\subfile{sections/6conclusions.tex}

\vspace*{-8pt}
\section{Acknowledgements}
\label{sec:acknowledgements}
\vspace*{-5pt}
\subfile{sections/7acknowledgements.tex}




\ninept
\label{sec:refs}
\bibliographystyle{IEEEbib}
\bibliography{main}

\end{document}

%% file: sections/0abstract.tex
An increasing number of datasets contain multiple views, such as video, sound and automatic captions. A basic challenge in representation learning is how to leverage multiple views to learn better representations. This is further complicated by the existence of a latent alignment between views, such as between speech and its transcription, and by the multitude of choices for the learning objective. We explore an advanced, correlation-based representation learning method on a 4-way parallel, multimodal dataset, and assess the quality of the learned representations on retrieval-based tasks. We show that the proposed approach produces rich representations that capture most of the information shared across views. Our best models for speech and textual modalities achieve retrieval rates from 70.7\% to 96.9\% on open-domain, user-generated instructional videos. This shows it is possible to learn reliable representations across disparate, unaligned and noisy modalities, and encourages using the proposed approach on larger datasets.

%% file: sections/1introduction.tex
%
Many large datasets include multiple modalities~\cite{lin2014microsoft}, leading to an exploration of methods that exploit the multimodal structure of the data. Further, collecting large datasets with multiple views is relatively easier than collecting large datasets with high quality annotations~\cite{abu-el-haija16}. Multiple views help learn better representations for each view separately \cite{ngiam11}, or a shared representation across multiple views~\cite{arora2013multi}, and multiview learning has also been shown to be useful in low-resource settings~\cite{socher10}. However, the fusion of information from disparate modalities remains a challenging problem \cite{baltruvsaitis18}.

\begin{figure}[ht]
\centering
     \includegraphics[width=0.46\textwidth]{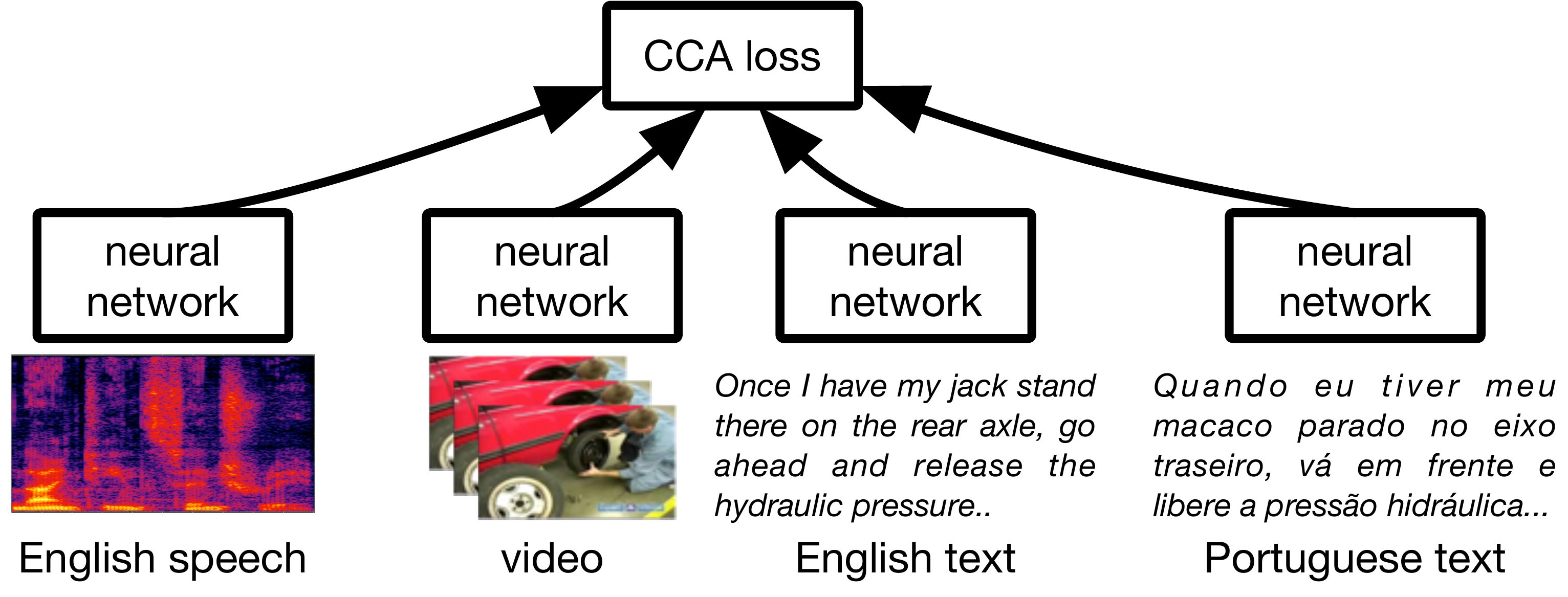}
      \caption{Learning from multiple modalities using DGCCA.}
       \label{fig:overview}
\end{figure}

In this paper, we build multiview models on a newly released multimodal dataset, the How2 corpus \cite{how2}, which contains user-generated instructional videos for a variety of tasks such as cooking, playing or dancing. While previous multiview models have exploited the natural alignment between views, such as speech and articulatory features \cite{wang2015unsupervised}, here we have to overcome challenges resulting from latently aligned views. For instance, there exists an alignment between words in an English sentence, and the words in its Portuguese translation. Figure \ref{fig:overview} shows an overview of our learning algorithm with 4 different input views, detailed in Section \ref{sec:methods}.

Given these multiple parallel modalities, we address the following: how much information is shared across modalities? How can we learn a representation that captures information from all modalities? We measure this using intrinsic evaluations. In the remainder of this paper, we describe related work, then the proposed methods and our experimental setup, and finally present our results.

%% file: sections/2related_work.tex


In audio-visual speech recognition, \cite{palaskar18, mroueh15} explore strategies to learn and fuse audio and visual representations in a neural net, including concatenating both modalities, bilinear products between representations, and weighted addition of modalities. Within a scene classification task, \cite{aytar18} use one neural network per modality. The weights of the last layers of each network are tied, and the distributions of representations are encouraged to be more similar across views. On most of the datasets considered, this improved the cross modal retrieval score. Using a parallel corpus for many languages, \cite{schwenk17} train language-specific RNN encoders and decoders on machine translation tasks, obtaining high cross-lingual retrieval scores on an out of domain corpus. Using a supervised end-to-end speech recognition model, \cite{palaskar2018acoustic} learn acoustic word embeddings and how to segment speech into words.


Representations can also be learned independently of a specific task, for instance with a hinge loss, where a neural network is trained to distinguish matched from mismatched pairs across or within modalities \cite{he16b}. Using this method, \cite{harwath17} learn representations for speech and images on a spoken caption dataset. Their method implicitly learns an alignment between speech segments and image regions, which the authors show correspond to word-like units. In a similar setting, \cite{arandjelovic2017objects} use triplet sampling and binary classification to learn feature extractors for audio and video data, and evaluate them on sound classification and image classification. \cite{chung18} learn representations for speech and text separately in an unsupervised way, then learn a mapping between both spaces that keeps both distributions undistinguishable for an adversarial net. 


Deep networks can learn representations by reconstructing one or two modalities, from one or two modalities \cite{ngiam11}. Better features for a single modality can be learned if multiple modalities are available at training time. A correlation-based loss is used in \cite{socher10} to learn representations for text and images on large unaligned corpora, which are then successfully used in a low-resource, supervised image captioning task. Reconstruction-based and correlation-based learning can be complementary, as described in \cite{wang15}, which examines both losses on an audio-visual speech recognition dataset. In \cite{yang17}, RNNs trained in a variety of supervised tasks are augmented with reconstruction losses, across- and within-modality, as well as a correlation-based loss.

%% file: sections/3methods.tex
In the following, we describe the sequence-to-sequence models we used to extract features for speech, text and video data. We then describe the correlation-based methods we used to learn representations.

\vspace*{-5pt}
\subsection{Input representations}
\label{ssec:input_representations}
\begin{figure}[ht]
\centering
     \includegraphics[width=0.49\textwidth]{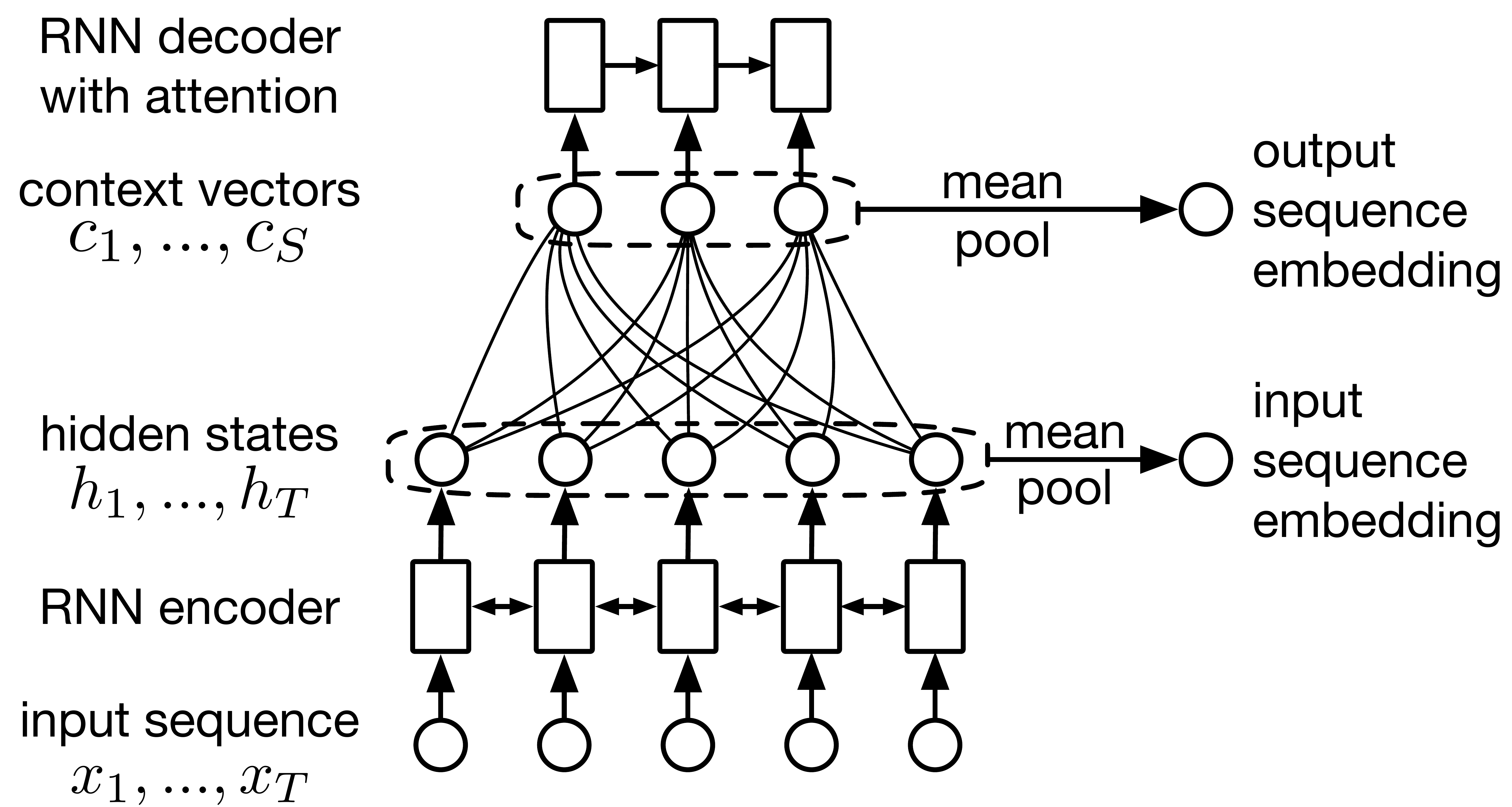}
      \caption{Extracting sequence embeddings from trained sequence to sequence models.}
       \label{sequence_embeddings}
\end{figure}

We use token-level representations from Machine Translation (MT) and Automatic Speech Recognition (ASR) systems to build sentence- or utterance-level representations. All the models we consider are attention-based, sequence to sequence models \cite{bahdanau14} that were trained in a supervised way on the How2 dataset. The encoder, a stacked bi-directional RNN, reads a sequence of feature vectors ${x_1, ..., x_T}$ and produces a sequence of hidden states ${h_1, ..., h_{T'}}$ (${T' \leq T}$ because of possible sub-sampling). The decoder, an RNN with attention mechanism, produces context vectors ${c_1, ..., c_S}$. We use the average of the ${h_i}$ (resp. ${c_i}$) as the representation for the input sequence (resp. output sequence), as depicted in Figure \ref{sequence_embeddings}. The RNNs for ASR (resp. MT) are LSTMs \cite{hochreiter97} (resp. GRUs \cite{cho14}). For the acoustic representations to be at the same level of granularity as the word representations from MT, we use the Acoustic2Word model \cite{palaskar2018acoustic} as our ASR model, and obtain acoustic embeddings $h_i$ at word level. 

For the video modality, we condense the information present in each utterance into a single vector as follows. We first use a ResNet~\cite{he16a} to map each frame of the video to a multi-class posterior, based on the 1000 ImageNet classes. We then compute the average of those posteriors. As the video frames are sampled with a hard temporal threshold, it may contain noisy artifacts. We average over all the frames to capture the most persistent predictions and reduce the variability due to noise. We experimented with representations from action networks \cite{hara3dcnns} trained on an action dataset \cite{kay2017kinetics}, and obtained similar results as with ResNet features.

\vspace*{-5pt}
\subsection{Linear CCA}
\label{subsec:lcca}



To measure how much information is shared by pairs of representations, we use Canonical Correlation Analysis (CCA) \cite{hotelling36}. We assume that we are given two views of the same data point: for instance, for a given utterance, the audio recording and the transcription. These two views are represented by random variables $X$ and $Y$ ($d_x$- and $d_y$-dimensional respectively). Linear CCA seeks two linear transformations $\U \in \Rm{d_x}{k}$ and $\V \in \Rm{d_y}{k}$ such that the components of $\U^TX$ and $\V^TY$ are maximally correlated. Formally, we want to maximize $\E{X,Y} [ \tr(\U^TXY^T\V) ]$ subject to ${\E{X} [ \U^TXX^T\U ] = \E{Y} [ \V^TYY^T\V ] = \I_k}$.

For dataset $\{x_i,y_i\}_{i=1}^N$, we define $\C_{XY}$ the empirical cross-covariance matrix between $X$ and $Y$, and $\C_{XX}$ and $\C_{YY}$ the empirical auto-covariance matrices of $X$ and $Y$, respectively. $\U$ and $\V$ are given by the $k$ left and right singular vectors of $\C_{XX}^{-1/2}\C_{XY}\C_{YY}^{-1/2}$ with the largest singular values.

CCA is a better objective than predicting one view with the other when no single regression provides a fully adequate solution. For instance, it is very hard to generate speech from text. Instead, it is easier to predict the dependent variate which has the largest multiple correlation.

\vspace*{-5pt}
\subsection{Extensions of CCA}
\label{subsec:dcca}

Deep CCA (DCCA) \cite{andrew2013deep} is a natural extension of linear CCA, where one seeks to maximally correlate $\U^Tf(X)$ and $\V^Tg(Y)$. $f$ and $g$ are non-linear feature extractors, which can be learned via gradient descent on the CCA objective. It is also natural to extend CCA to multiple views \cite{horst61}. 

Instead of 2 views, we have $J$ views ${X_1, ..., X_J}$ attached to each data point, stored in matrices $\X_j \in \Rm{d_j}{N}$. One finds linear transformations $\{\U_j \in \Rm{d_j}{k}\}_{j=1}^J$, that minimize the mutual reconstruction error under constraints, in a way equivalent to maximizing correlation. This framework can be extended to non-linear feature extractors \cite{benton17} with the objective:
\linebreak
minimize $\sum \limits_{j=1}^{J}||\G - \U_j^Tf_j(\X_j)||_2^2$ subject to ${\G\G^T =\I_k}$, with respect to parameters $\{\G, \{f_j, \U_j\}_j\}$. Here, ${f_j(\X_j)\in\Rm{h_j}{N}}$ is the output of $j$-th feature extractor, and ${\G \in \Rm{k}{N}}$ can be viewed as the learned representations for the dataset. Effectively, each pair of feature extractor $f_j$ and linear transformation $\U_j$ tries to reconstruct the learned representation $\G$. The constraints on $\G$ prevent the feature extractors from collapsing the features. We refer to this method as deep generalized CCA (DGCCA). A linear variant of DGCCA where the $f_j$'s are identity maps has been applied to acoustic feature learning~\cite{arora2014multi} and learning word embeddings~\cite{rastogi2015multiview}.

%% file: sections/4experiments.tex
We applied the methods described above to a 4-way parallel corpus, and evaluated the learned representations using a retrieval task.

\vspace*{-5pt}
\subsection{Dataset}
\label{ssec:dataset}

We apply the described methods to the How2 dataset \cite{how2}, which we use as a 4-way parallel corpus: video, speech, transcription in English, translation in Portuguese. The dataset contains 13,500 videos, or 300 hours of speech, and is split into 185,187 training, 2022 development (dev), and 2361 test utterances. 

It is yet unclear how much temporal coherence there is between the video modality and the language (text and speech) content, as objects mentioned early in the video may only appear much later on screen, or only for a very brief time. 

Whenever we mention an MT task, it consists of translating English (en) text to Portuguese (pt) text; an ASR task consists of transcribing English speech to English text; a Speech Translation (ST) task consists of mapping English speech to Portuguese text.

\vspace*{-5pt}
\subsection{Evaluation}
\label{ssec:evaluation}

To measure the richness of the learned representations, we use them in a retrieval task: given a source sequence in view~1, and a set of reference sequences in view~2, find the $n$ sequences in the reference set that are closest to the source sequence. Since we have parallel corpora, we can check whether the correct sequence is present within those $n$ sequences. We report this as Recall@10 (${n=10}$ throughout this paper), with scores ranging from 0 to 100. Finding the closest sequences consists of projecting the reference set as well as the source sequence into the shared space, then computing distances between source and references, and retrieving the closest points. The Recall@10 of picking at random from the reference set is $0.5\%$ for the dev set and $0.4\%$ for the test set.

%% file: sections/5results_discussion.tex
In the following, we will use $k$ to indicate the dimensionality of the shared representation. Typically, $k$ should be at most the smallest dimensionality of all views involved. We set $k$ to half the smallest dimensionality as a balance between keeping as much information as possible while dropping uninformative components. Throughout all our experiments, we add the identity matrix scaled by $10^{-16}$ to the view-specific co-variance matrices. 

In all experiments involving DCCA and DGCCA, we use 2-layer feedforward neural networks as feature extractors ($f, g, f_j$ in Section \ref{subsec:dcca}), the first layer with the same size as the input, and the second of size $k$. The training proceeds in epochs, which consist of a full pass over the training set with batch size 5500. After each epoch, we compute the retrieval scores between all possible pairs of different views on the dev set, and aggregate the scores by picking the highest of those scores. Our final model is the one with the highest aggregate score. For the experiments involving the video modality, we used a weight decay of $10^{-5}$.

\vspace*{-5pt}
\subsection{Bimodal Experiments}
\label{subsec:bimodal}

We start by applying linear CCA and deep CCA to pairs of views, at the utterance level. Text, speech and video sequences are represented with 800-, 320- and 1000-dimensional vectors respectively. As measured by the retrieval rates shown in Table \ref{tab:bimodal_retrieval}, the representations learned for text (en and pt) and speech capture almost all of the information present in both views, in a space with half the dimensionality. The original text (en) and text (pt) representations having the same dimensionality, we scored the retrieval of a Portuguese sentence given an English sentence, which yielded a score of 0.38\%. For other pairs of modalities, there is no obvious way of computing pairwise distances in the original space. The retrieval scores involving the video modalitiy are very low, and we discuss those in Section~\ref{subsec:discussion}. 

\begin{table}[h!]
\caption{Recall@10 for retrieving reference modality given source modality ("source - reference"). Swapping source and reference change retrieval scores by less than 1\% absolute.}
\centering
\begin{tabular}{ lrrrrr }
\hline
\hspace*{0.2cm}
& \multicolumn{2}{c}{Linear CCA} & \multicolumn{2}{c}{Deep CCA} & \\ 
\cmidrule(r){2-3}\cmidrule(r){4-5}
& dev & test & dev & test & $k$\\
\hline
text (en) - text (pt) & 82.5 & 81.4      & 95.1 & 94.6      & 400 \\
speech - text (en)    & 98.3 & 96.9      & 92.1 & 90.1      & 160 \\
video - text (en)     & 0.9 & 0.8       & 2.32 & 1.6       & 400 \\
video - speech        & 0.8 & 0.6       & 1.93 & 1.8       & 160 \\
\hline
\end{tabular}
\label{tab:bimodal_retrieval}
\end{table}

To tie the results in Table~\ref{tab:bimodal_retrieval} to known metrics, we take the first retrieval result and score it as though it were the output of an ASR or MT system. Given a speech utterance from the test set that we want to transcribe, or a source sentence we wish to translate, we pick the closest sentence from a reference set using our learned DCCA model. We then score this pick using the relevant metric, WER for ASR and BLEU for MT. The score strongly depends on the contents of the reference set: if the reference set contains no appropriate sentence to transcribe (resp. translate) the source, the WER (resp. BLEU) will be high (resp. low). We thus test on two reference sets: 1) the training set, 2) the union of the training and test set. In setting 1, the reference set does not contain the correct answers, whereas it does in setting 2. When using only the test set as a reference set, the score is almost perfect, and we only report the more challenging settings, in columns WER and BLEU (MT) of Table \ref{tab:scoring_retrieval}. The results on the train set are quite poor given that the train set may not contain appropriate sentences for the test set. We estimate this by finding, for each sentence in the test set, the closest sentence (in terms of edit distance) from the train set. This yields a BLEU of 10.6, and a WER of 63.0\%. When using the union of test and train as a reference set, our model is still able to mostly pick out the correct sentences, achieving on par or better performance than a baseline sequence-to-sequence model. This is consistent with our retrieval scores, as the retrieval for text and text was slightly higher than speech and text.

\begin{table}[h!]
\caption{Scoring top-1 retrieval result from DGCCA models with ASR, MT and ST metrics. Models used (from left to right) were trained using speech and text (en); text (en) and text (pt); speech, text (en), text (pt) and video. Source sentences for the retrieval are from the test set.}
\vspace*{-5pt}
\centering
\begin{tabular}{ llll }
\hline
\hspace*{0.2cm}
\textbf{Reference Set}  & WER  & BLEU (MT) & BLEU (ST) \\ 
			        \hline
train               & 134\% & 5.2 & 0.2\\
train + test        & 27.4\% & 80.7 & 19.8 \\
Baseline S2S        & 24.3\% & 57.3 & 27.9 \\
                    \hline
\end{tabular}
\label{tab:scoring_retrieval}
\end{table}

\vspace*{-9pt}
\subsection{$n$-modal experiments}
\label{subsec:nmodal}

In subsequent experiments, we used DGCCA to learn representations with more than 2 views. We learned representations with English text, speech and video, with ${k=160}$, and report retrieval results in Table \ref{tab:speech_text_video}. As compared to Table \ref{tab:bimodal_retrieval}, the retrieval scores between speech and text (en) decrease, as the model has to accommodate a third view. Keeping hyperparameters fixed and adding a fourth view, Portuguese text, we obtain the results in Table \ref{tab:speech_text_text_video}. Relative to Table \ref{tab:bimodal_retrieval}, the text - text retrieval score increases, while the speech - text (en) score decreases, and the scores involving video decrease slightly. This could be explained by the fact that text - text retrieval is an easier task than those involving speech and video, so that the model trades off a higher loss in the video and speech domain for a lower loss in the text domain. To remedy this, one could add weights $w_j$ to each reconstruction loss in Section \ref{subsec:dcca}, or tune the architectures of the $f_j$. As in Section \ref{subsec:bimodal}, we evaluate our speech - text (pt) retrieval with an ST task. The results are shown in column BLEU (ST) of Table \ref{tab:scoring_retrieval}, and are again consistent with the retrieval scores.

\begin{table}[h!]
\caption{Recall@10 for retrieving column modality given source row modality, for a DGCCA model trained on 3 views as described in Section~\ref{subsec:nmodal}. Results from the bottom left triangle can be compared to those in Table \ref{tab:bimodal_retrieval}.}
\vspace*{-5pt}
\centering
\begin{tabular}{ lllll }
\hline
\hspace*{0.2cm}
& & text (en) & speech & video \\ 
			        \hline
text (en) & dev & - & 92.1 & 1.7 \\
 & test & - & 89.8 & 1.8 \\
speech & dev & 92.1 & - & 1.9 \\
 & test & 89.1 & - & 1.2 \\
video & dev & 1.4 & 1.9 & - \\
 & test & 1.7 & 1.2 & - \\
                    \hline
\end{tabular}
\label{tab:speech_text_video}
\end{table}

\begin{table}[h!]
\caption{Recall@10 for retrieving column modality given source row modality, for a DGCCA model trained on 4 views as described in \ref{subsec:nmodal}. Results from the bottom left triangle can be compared to those in Table \ref{tab:bimodal_retrieval}.}
\vspace*{-5pt}
\centering
\begin{tabular}{ llllll }
\hline
\hspace*{0.2cm}
          &      & Text (pt) & Text (en) & Speech & Video \\ 
			        \hline
Text (pt) & dev  & -         & 98.8      & 73.5   & 2.1   \\
          & test & -         & 98.3      & 71.0   & 1.1   \\
Text (en) & dev  & 98.8      & -         & 88.2   & 1.4   \\
          & test & 98.4      & -         & 85.4   & 0.9   \\
Speech    & dev  & 73.0      & 88.1      & -      & 1.1   \\
          & test & 70.7      & 85.4      &-       & 1.0   \\
Video     & dev  & 2.1       & 1.1       & 1.0    & -     \\
          & test & 1.1       & 1.1       & 0.9    & -     \\
                    \hline
\end{tabular}
\label{tab:speech_text_text_video}
\end{table}

\vspace*{-5pt}
\subsection{Discussion}
\label{subsec:discussion}

As shown by our Recall@10 retrieval results, the CCA objective induces a shared space capturing most of the information shared across the original spaces. Scoring the top-1 retrieved data point with common MT, ASR and ST metrics is consistent with this finding. Moreover, this shared space is learned on top of high-level, unrelated representations: the training of the ASR and MT systems is entirely independent.

Our results involving video are not in agreement with that hypothesis, and we see two possible explanations. First, there is a temporal mismatch between the video modality and the language content, as described in Section \ref{ssec:dataset}. Second, it is possible that the ResNet posteriors are either extremely noisy, or simply fail to identify certain relevant objects because of a domain mismatch, as discussed in Section \ref{ssec:input_representations}. Previous work in the context of ASR shows that using the penultimate instead of the last layer of the ResNet makes little difference \cite{palaskar18}.

%% file: sections/6conclusions.tex
In this paper, we cast the How2 dataset to a multiview, multimodal representation learning problem. We explore an advanced, correlation-based learning technique, for two or more views, and evaluate the learned representations using cross-view retrieval tasks. Our results show that the geometry of the embedding space captures the information necessary to relate data points of various modalities, with a dimensionality much smaller than that of the original spaces.

Overall, our retrieval results indicate that downstream tasks may benefit from integrating these learned representations. This must be implemented very carefully and is left for future work.

%% file: sections/7acknowledgements.tex

The work reported here was conducted at JSALT 2018, and supported by JHU with unrestricted gifts from Amazon, Facebook, Google, Microsoft and Mitsubishi Electric Research Laboratories. This work was also supported in part by NSF BIGDATA grant IIS-1546482, H2020 ERC Starting Grant No. 678017, Newton Fund Institutional Links Grant ID 352343575, NSF grant number OCI-1053575, and NSF award number ACI-1445606.  We thank Adrian Benton for fruitful discussions and guidance with DGCCA, and the JSALT 2018 team for helpful feedback.